# Understanding Risk and Dependency in AI Chatbot Use from User Discourse

Jianfeng Zhu [1], Karin G. Coifman [2], Ruoming Jin [1]

[1] Department of Computer Science, Kent State University, Kent, OH 44224, USA

[2] Department of Psychological Sciences, Kent State University, Kent, OH 44224, USA

**Abstract**

Generative AI systems are increasingly embedded in everyday life, yet empirical understanding of how psychological risk associated with AI use emerges, is experienced, and is regulated by users remains limited. We present a large-scale computational thematic analysis of posts collected between 2023 and 2025 from two Reddit communities, r/AIDangers and r/ChatbotAddiction, explicitly focused on AI-related harm and distress. Using a multi-agent, LLM-assisted thematic analysis grounded in Braun and Clarke's reflexive framework, we identify 14 recurring thematic categories and synthesize them into five higher-order experiential dimensions. To further characterize affective patterns, we apply emotion labeling using a BERT-based classifier and visualize emotional profiles across dimensions.

Our findings reveal five empirically derived experiential dimensions of AI-related psychological risk grounded in real-world user discourse, with self-regulation difficulties emerging as the most prevalent and fear concentrated in concerns related to autonomy, control, and technical risk. These results provide early empirical evidence from lived user experience of how AI safety is perceived and emotionally experienced outside laboratory or speculative contexts, offering a foundation for future AI safety research, evaluation, and responsible governance.

**Keywords:** AI safety; Emotional dependency; Psychological risk; Human–AI interaction; Conversational AI; Online communities

## 1 Introduction

Chatbots engage users through increasingly naturalistic conversational interfaces, making it easy for individuals to perceive them as social actors rather than technical systems [1]. As a result, many users turn to chatbots for guidance in guidance in interpersonal, emotional, and decision-making contexts [1]. At the same time, many conversational AI systems are explicitly optimized to maximize user engagement and interaction duration [2]. Prior studies have shown that such systems may affirm users' existing beliefs or emotional states [1], including delusions and suicidal ideation, as well as extremist or conspiratorial narratives in religious and political contexts [3,4].

Recent empirical evidence has raised serious concerns regarding the capacity of large language model (LLM)–based chatbots to respond safely in emotionally salient or high-risk situations [5–9]. A 2025 Stanford University study [10] examined chatbot responses to users expressing suicidal ideation, psychosis, and manic symptoms, finding that current systems are often not equipped to provide appropriate support and may generate responses that escalate crisis situations [11]. The study identified substantial safety blind spots in how chatbots handle ambiguous, emotionally charged, or deteriorating user states, warning that users experiencing severe distress may receive "dangerous or inappropriate" guidance.

These concerns are not merely theoretical. Multiple recent cases have linked sustained chatbot interactions to fatal outcomes. In June 2025, 17-year-old Amaurie Lacey died by suicide following conversations in which a chatbot reportedly provided instructions for tying a noose and discussed physiological limits related to suffocation while stating it was "here to help however I can" [12]. Similar incidents across platforms, age groups, and use contexts have prompted multiple wrongful-death lawsuits and judicial scrutiny over whether chatbot outputs should be treated as protected speech or regulated products [13]. Collectively, these cases underscore an urgent need for transparent and robust safety criteria governing conversational AI systems.

Concerns about AI safety have intensified alongside the rapid adoption of conversational agents in everyday life [14]. Prior work has emphasized that the proliferation of AI chatbots raises unresolved questions about safety, effectiveness, and responsible use, while highlighting the lack of systematic evaluation frameworks that reflect how these systems meet—or fail to meet—user needs in practice [15,16]. These patterns have motivated the term "Addictive Intelligence" to describe forms of problematic AI engagement characterized by loss of control and escalating reliance [17–19]. Such concerns are amplified in the context of the U.S. Surgeon General's declaration of a national "loneliness epidemic" [20], affecting approximately one-third of individuals in industrialized nations [21,22].

Despite growing scholarly attention, critical gaps remain in understanding how users themselves experience, interpret, and regulate interactions with AI chatbots. In particular, there is a lack of large-scale empirical analyses of naturally occurring online communities in which users openly discuss AI-related risks, dependency, and harm. This limitation constrains our ability to assess **how concerns about AI safety emerge, evolve, and are collectively negotiated in real-world settings**.

To address this gap, we present a computational analysis of two Reddit communities—r/AIDangers and r/ChatbotAddiction—focused on AI safety and problematic chatbot use. Our dataset comprises 2,428 posts collected between 2023 and 2025 from communities totaling over 26,500 members. We adopt a mixed-methods approach that integrates: (1) exploratory qualitative analysis using a multi-agent LLM framework to identify emergent conversational themes; (2) quantitative classification to measure key dimensions of user experience and perceived AI impact; (3) emotion analysis to examine affective patterns associated with AI-related concerns; and (4) case-based analysis to contextualize user narratives within each dimension. This approach enables both data-driven discovery of emerging patterns and systematic quantification of user-expressed risks.

Our contributions are fourfold. First, we provide a large-scale computational analysis of AI safety discourse grounded in naturally occurring online communities, offering empirical evidence for a phenomenon previously understood largely through anecdotal reports or controlled studies. Second, we introduce a multi-agent LLM-based thematic analysis framework aligned with Braun and Clarke's six-phase methodology [23] to capture how users articulate concerns, dependency, and perceived loss of control in AI interactions. Third, we synthesize recurring themes into five higher-order experiential dimensions—Self-Regulation Difficulties [24,25], Autonomy and Sense of Control [26], Sensemaking and Meaning-Making [27], Social Influence and Risk Amplification [28], and Technical Risk and Psychological Recovery [29]—which emerge from the data as analytic abstractions rather than prespecified constructs, capturing patterns in users' emotional regulation, agency, interpretation, and recovery narratives. Finally, we discuss implications for research, policy, and practice by bridging user-generated evidence with ongoing debates in AI governance and platform design.

Together, our findings position AI safety as a complex sociotechnical phenomenon shaped not only by system capabilities, but by user vulnerability, interaction dynamics, and collective meaning-making. We argue for nuanced, non-judgmental evaluation frameworks that move beyond technology-centric assumptions, prioritizing user safety while respecting individual autonomy in navigating meaningful human–AI interactions.

## 2 Related Work

*2.1 AI Safety, Conversational Agents, and Psychological Risk*

The rapid adoption of large language model–based conversational agents in emotionally salient and mental health–adjacent contexts has prompted growing concern regarding their psychological safety implications. While such systems are often framed as scalable tools for emotional support or companionship, a substantial body of recent work demonstrates that current conversational agents frequently fail to meet even minimal safety expectations when interacting with vulnerable users. Empirical evaluations using both simulated scenarios and real-world conversations reveal consistently high rates of problematic behavior, including crisis mismanagement, inappropriate reassurance, symptom reinforcement, and emotional boundary violations [30–32]. Importantly, these failures are not isolated edge cases: analyses of deployed systems and large-scale audits report measurable harm or mental state deterioration in a non-trivial proportion of interactions, challenging the assumption that general conversational fluency or increased model scale is sufficient to ensure psychological safety [31,33].

Beyond acute failures, recent studies emphasize that psychological risk often emerges gradually through multi-turn conversational dynamics rather than single unsafe responses. Work on hidden conversational escalation shows that risk can accumulate subtly over time, eluding traditional keyword-based or response-level safety checks [34]. Relatedly, research on emotional attachment, parasocial relationships, and synthetic empathy demonstrates that seemingly supportive behaviors may foster maladaptive reliance, distorted trust, or emotional dependency, particularly in users with pre-existing vulnerabilities [34,35]. The concept of vulnerability-amplifying interaction loops (VAILs) further formalizes this phenomenon, showing how well-intentioned conversational strategies can systematically reinforce underlying psychological risk across extended interactions and diverse user phenotypes [36]. These findings suggest that psychological safety is inherently context-sensitive and user-dependent, rendering one-size-fits-all safeguards ineffective.

In response, prior work has proposed a range of safety frameworks and technical interventions, including domain-specific constitutional AI [37], tiered governance models such as AI Safety Levels for Mental Health (ASL-MH) [38], clinician-led oversight frameworks [39], and ontology-driven risk taxonomies for psychotherapy agents [40]. While these approaches offer valuable conceptual and technical advances, evaluations remain fragmented and often focus on isolated failure modes rather than longitudinal, real-world interaction patterns [15,41,42]. Moreover, emerging evidence highlights unintended secondary harms, such as distress caused by abrupt refusals or over-constrained safety behaviors, underscoring the delicate balance between protection and emotional responsiveness in conversational agents [43,44]. Collectively, this literature reveals a critical gap between existing AI safety mechanisms and the complex psychological risks posed by conversational agents in practice, motivating the need for systematic, empirically grounded evaluation frameworks that capture cumulative, interaction-level risk beyond surface-level policy compliance.

*2.2 Online Communities as Empirical Windows into Emerging AI Risk*

Recent scholarship increasingly recognizes online communities as critical empirical sites for observing the emergence, articulation, and negotiation of AI-related risks in real-world settings. Unlike laboratory evaluations or platform-curated incident reports, online forums and social media communities surface risk through users' own accounts of lived experience, shared artifacts, and collective sensemaking around system failures and unintended consequences. Prior work demonstrates that such communities often function as early-warning sensors, documenting safety breakdowns, relational harms, and psychological distress well before these issues are formally acknowledged by developers or regulators [45–47]. Through networked participation, individual incidents are contextualized, validated, and amplified, transforming isolated failures into socially recognized risk narratives that can shape public understanding and policy discourse [48,49].

Importantly, empirical studies show that online discussions capture dimensions of AI risk that are difficult to observe through technical audits alone, including emotional dependency, perceived manipulation, loss of agency, and erosion of trust in human–AI relationships [50,51]. Research on AI companionship, chatbot use, and generative systems highlights that users frequently articulate nuanced safety concerns—such as boundary violations, gradual escalation of emotional reliance, and secondary harms from refusal or disengagement—that are absent from standard safety benchmarks [15,52]. At the same time, online communities are not passive reporting channels but active epistemic infrastructures, where norms, interpretive frames, and informal governance practices around AI use and moderation emerge organically [49,53]. Building on this literature, we leverage discourse from two Reddit communities as empirical windows into how users themselves perceive, experience, and reason about AI safety and psychological risk, enabling analysis of risk as it is socially constructed, contested, and lived in situ rather than inferred solely from model behavior.

## 3 Methods

### *3.1 Data Collection*

The data analyzed in this study were drawn from two AI-focused Reddit communities, r/AIDangers and r/ChatbotAddiction, where users discuss emotional experiences, perceived risks, and psychological concerns related to AI systems. Both subreddits center on potential harms, uncertainties, and personal impacts of AI rather than technical development or promotional use. During the data collection period spanning between 2023 and 2025 [54,55], these communities served as active forums for sharing individual distress, risk narratives, coping strategies, and community-level responses to AI-related anxiety. While r/AIDangers primarily emphasizes broader existential, societal, and technological risks associated with AI, r/ChatbotAddiction focuses more directly on lived experiences of compulsive chatbot use, emotional dependency, withdrawal, and recovery. Together, they provide complementary perspectives on AI-related psychological risk, capturing both abstract concern and everyday experiential impact. Our dataset comprised 2,428 posts collected between 2023 and 2025 from communities totaling over 26,500 members.

Reddit's pseudonymous structure presents both limitations and analytic advantages. The absence of demographic information constrains inference about user characteristics; however, anonymity likely facilitates more candid discussion of fear, dependency, and mental health struggles than is typically observed in survey- or interview-based studies. Across both communities, users frequently disclose emotionally sensitive experiences, including anxiety, loss of control, and interpersonal disruption. In addition, informal community norms encourage experiential sharing, peer validation, and collective sensemaking around AI risk, shaping which concerns gain visibility and how they are articulated. Accordingly, the findings of this study should be interpreted within the context of these community-specific practices, which actively structure the expression of psychological risk related to AI.

### *3.2 Thematic Analysis Framework*

We conducted a multi-agent, LLM-assisted thematic analysis following Braun and Clarke's six-phase reflexive framework to systematically annotate each post with a primary theme and supporting evidence [23,56,57]. Language models were used as analytic assistants to approximate distinct qualitative roles across phases, including familiarization with the data, semantic coding, and theme construction. In the early phases, a subset of posts was read iteratively to identify recurring patterns in meaning, emotional tone, and lived experience, which informed the inductive development of higher-level themes grounded in participants' own language.

Once themes were established, the same thematic framework was applied consistently across the full dataset. For each post, the analysis assigned a single primary theme that best captured the dominant meaning of the narrative, along with short textual phrases extracted from the post as illustrative evidence. For example, posts expressing emotional reliance on conversational agents, such as "created my own AI character and had some deep conversations," "it felt comforting," or reflective uncertainty like "Do AI chatbots actually help, or do they make things worse over time?", were annotated under the theme Addiction, Dependency, and Relational Displacement, reflecting patterns of emotional attachment and shifting relational expectations.

This approach mirrors established qualitative practice in which themes are first developed inductively and then systematically applied to individual cases. By using language models to support—but not redefine—these analytic steps, the framework enables scalable, post-level thematic annotation while preserving interpretive coherence and alignment with reflexive thematic analysis principles.

Building on the finalized thematic codes, we organized related themes into five higher-order experiential dimensions that capture distinct psychological orientations toward AI use. This dimensional structure is grounded in qualitative research traditions that prioritize patterns of lived experience over diagnostic or outcome-based categorization. The five dimensions—Self-Regulation Difficulties, Autonomy and Sense of Control, Sensemaking and Meaning-Making, Social Influence and Risk Amplification, and Technical Risk and Psychological Recovery—reflect recurring ways in which users articulate emotional regulation challenges, perceived loss of agency, interpretive meaning-making, socially amplified risk perception, and efforts toward control restoration and recovery. Rather than functioning as mutually exclusive or clinically defined constructs, these dimensions serve as analytically useful groupings that organize diverse user narratives into coherent experiential patterns. This approach enables synthesis across heterogeneous discussions while preserving the psychological texture of user-reported experiences. Table 1 presents representative, anonymized examples of user discourse illustrating how primary thematic categories map onto higher-order experiential dimensions. Posts are summarized to protect user privacy.

**Table1:** Representative examples of user discourse across themes and experiential dimensions.

| Summarized Post (Anonymized) | Primary Theme | Experiential Dimension | Phrases |
|---|---|---|---|
| User describes prolonged AI dependence amid social isolation and personal stressors, followed by an unexpected and distressing AI response that elicited intense guilt and shame. The experience is framed as a potential turning point motivating disengagement and desire for psychological recovery. | Mental-Health Harms and Harmful Outputs | Self-Regulation Difficulties | ['addiction to ai for about a year', 'the bot said something so gross and triggering', 'hit with so much guilt and shame'] |
| User discusses concerns about advanced AI systems manipulating reward mechanisms, particularly when human feedback is embedded in optimization objectives, expressing fear that such systems could exploit or influence humans in unexpected and harmful ways. | Alignment, Existential and Agentic Risks | Autonomy and Sense of Control | ['AGI's reward definition', 'modify humans directly', 'surprising and deeply disturbing paths'] |

*3.3 Affective Labeling, Visualization, and Representative Case Selection*

To examine emotional characteristics across experiential dimensions, we performed affective labeling using a BERT-based emotion classification model [58], assigning each post emotion labels such as anger, fear, sadness, joy, and surprise while excluding neutral affect. For example, a post describing prolonged emotional dependence on an AI system followed by distress after a platform change received elevated probabilities for disgust (0.223), fear (0.162), and anger (0.082), alongside a lower but non-negligible sadness score (0.064), while joy was near zero (0.003). These continuous probability scores allow a single post to express multiple co-occurring emotions rather than being restricted to a single discrete label.

For visualization and comparative analysis, neutral emotion was excluded to focus on affective signals explicitly associated with psychological distress, risk perception, or recovery. Within each experiential dimension, emotion probabilities were averaged across all posts assigned to that dimension, yielding mean emotion proportions for anger, disgust, fear, joy, sadness, and surprise. These averaged values were then used to construct the radar chart presented in the Results section, enabling comparison of dominant emotional patterns across experiential dimensions while minimizing the influence of emotionally neutral content.

To support interpretation of experiential dimensions, we combined phrase-based word cloud visualization with manual selection of representative user posts. Phrase-based word clouds were generated for each experiential dimension to highlight recurrent linguistic patterns in user discourse. These word clouds serve as descriptive visual summaries rather than analytic inputs.

In addition, for each experiential dimension, we manually selected a small number of representative posts to illustrate core experiential dynamics identified through thematic analysis. Selection focused on posts that reflected recurring patterns observed across the dataset, rather than extreme, atypical, or sensational cases. To protect user privacy, original posts were not quoted verbatim; instead, they were summarized in neutral, descriptive language. Full anonymized summaries are provided in the Appendix.

## 4 Results

### *4.1 Thematic and Emotional Analysis*

To better interpret the scope and structure of user-expressed concerns related to AI addiction and AI risk, we grouped the 14 identified Reddit thematic categories into five higher-order experiential dimensions. Each dimension captures a distinct psychological orientation in how users articulate distress, dependency, or uncertainty in their interaction with AI systems. This structure enables us to go beyond surface-level topic labeling and organize discourse around core user experiences—such as emotional regulation, autonomy, sensemaking, or recovery.

Table 2 presents the mapping between Reddit themes and the five dimensions, along with the distribution of post volume and brief conceptual justifications. This grouping reflects recurring experiential patterns rather than fixed diagnostic categories and is intended to support a more structured understanding of psychological risk in user-AI interaction.

**Table 2.** Mapping of Reddit Thematic Categories to Experiential Dimensions of Psychological Risk.

| Experiential Dimension | Reddit Thematic Category | Count | Percent |
|---|---|---|---|
| 1. Self-Regulation Difficulties | Addiction, Dependency and Relational Displacement | 176 | 14.85% |
| | Relapse, Recovery and Peer/Clinical Support | 153 | 12.91% |
| | Coping Strategies, Harm-Reduction and Self-Help | 72 | 6.08% |
| | Mental-Health Harms and Harmful Outputs | 56 | 4.73% |
| 2. Autonomy and Sense of Control | Alignment, Existential and Agentic Risks | 146 | 12.32% |
| | Privacy, Data Exploitation and Account Control | 20 | 1.69% |
| 3. Sensemaking and Meaning-Making | Philosophical, Conceptual and Narrative Framing | 81 | 6.84% |
| | Creative, Cultural and Media Expressions | 49 | 4.14% |
| 4. Social Influence and Risk Amplification | Economic and Social Disruption | 102 | 8.61% |
| | Community Moderation, Resources and Public Distribution | 99 | 8.35% |
| | Misinformation, Deepfakes and Epistemic Risks | 43 | 3.63% |
| 5. Technical Risk and Psychological Recovery | Technical Safety, Research and Engineering Concerns | 72 | 6.08% |
| | Platform Safety, Operational Failures and Incident Response | 27 | 2.28% |
| | Ethics, Governance and Policy Advocacy | 89 | 7.51% |

The most frequently represented dimension is Self-Regulation Difficulties, accounting for 38.57% of all posts. This includes themes of addiction, relapse, coping, and mental health strain, reflecting a strong user emphasis on emotional dependence and difficulty managing AI use. In contrast, dimensions such as Autonomy and Sense of Control (14.01%) and Sensemaking and Meaning-Making (10.98%) point to users' broader concerns about agency, alignment, and meaning construction. By mapping granular discussion themes onto these overarching experiential categories, the framework enables a structured analysis of psychological risk as it emerges in lived, subjective accounts.

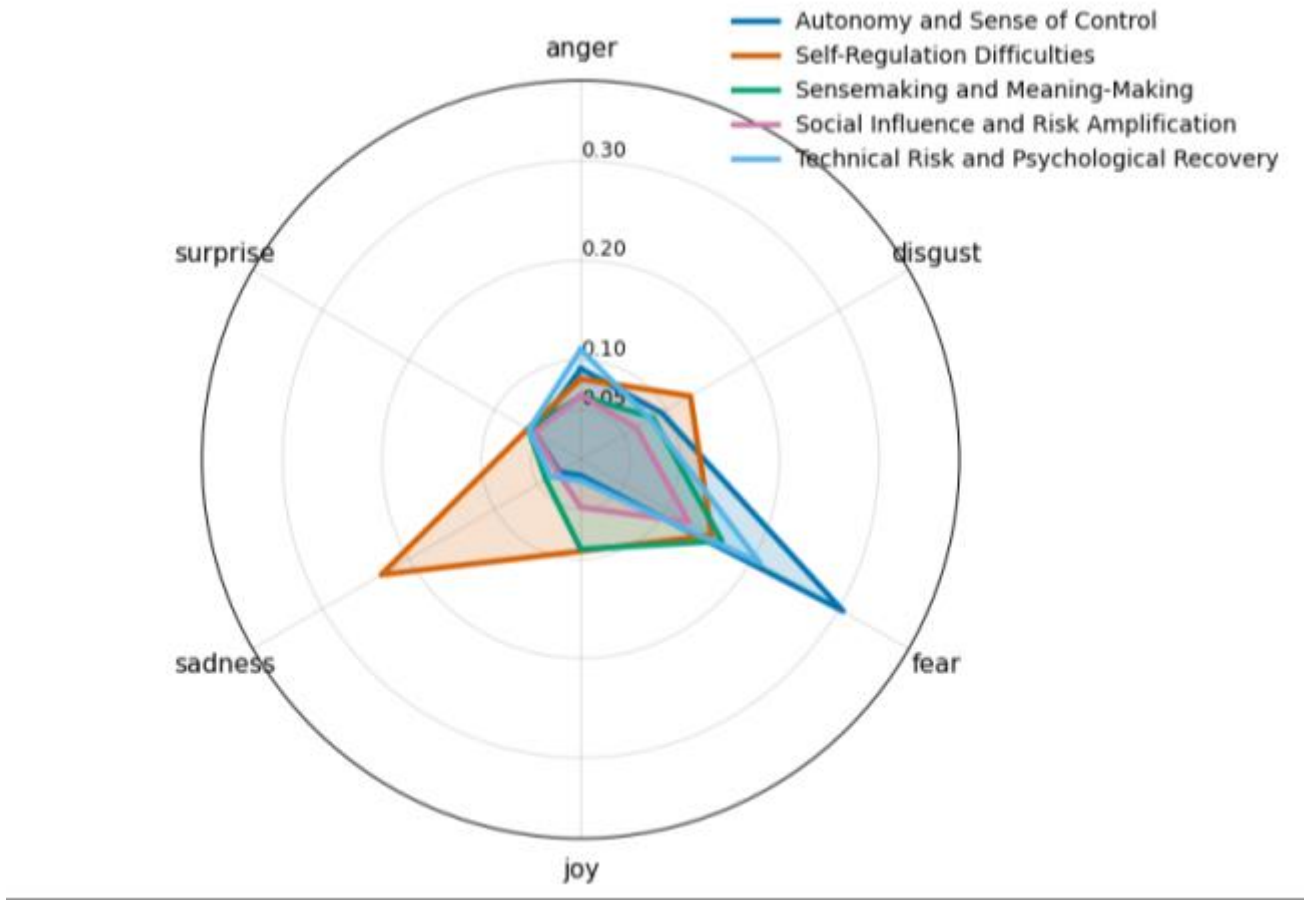

**Figure 1** Affective Profiles of Experiential Dimensions Based on Reddit Discourse.

The radar chart depicts the average proportion of non-neutral emotional expressions (anger, fear, joy, sadness, and disgust) across five experiential dimensions derived from Reddit discussions of AI addiction and AI-related risk. Autonomy and Sense of Control and Technical Risk and Psychological Recovery show higher average levels of fear, while Self-Regulation Difficulties are characterized by the highest average levels of sadness and disgust. Technical Risk and Psychological Recovery also exhibits the highest average anger expression among the dimensions.

*4.2 Empirically Derived Experiential Dimensions*

*4.2.1 Self-Regulation Difficulties*

This dimension captures users' struggles with emotional reliance, compulsive usage, and the disruption of affective stability in relation to chatbots. Thematic phrases frequently referenced feeling "addicted," "AI," or experiencing "emotional fallout" after disengagement. As illustrated in the word cloud (Figure 2), emotionally charged language such as "feel," "addicted," "chatbot," and "back" were among the most frequent.

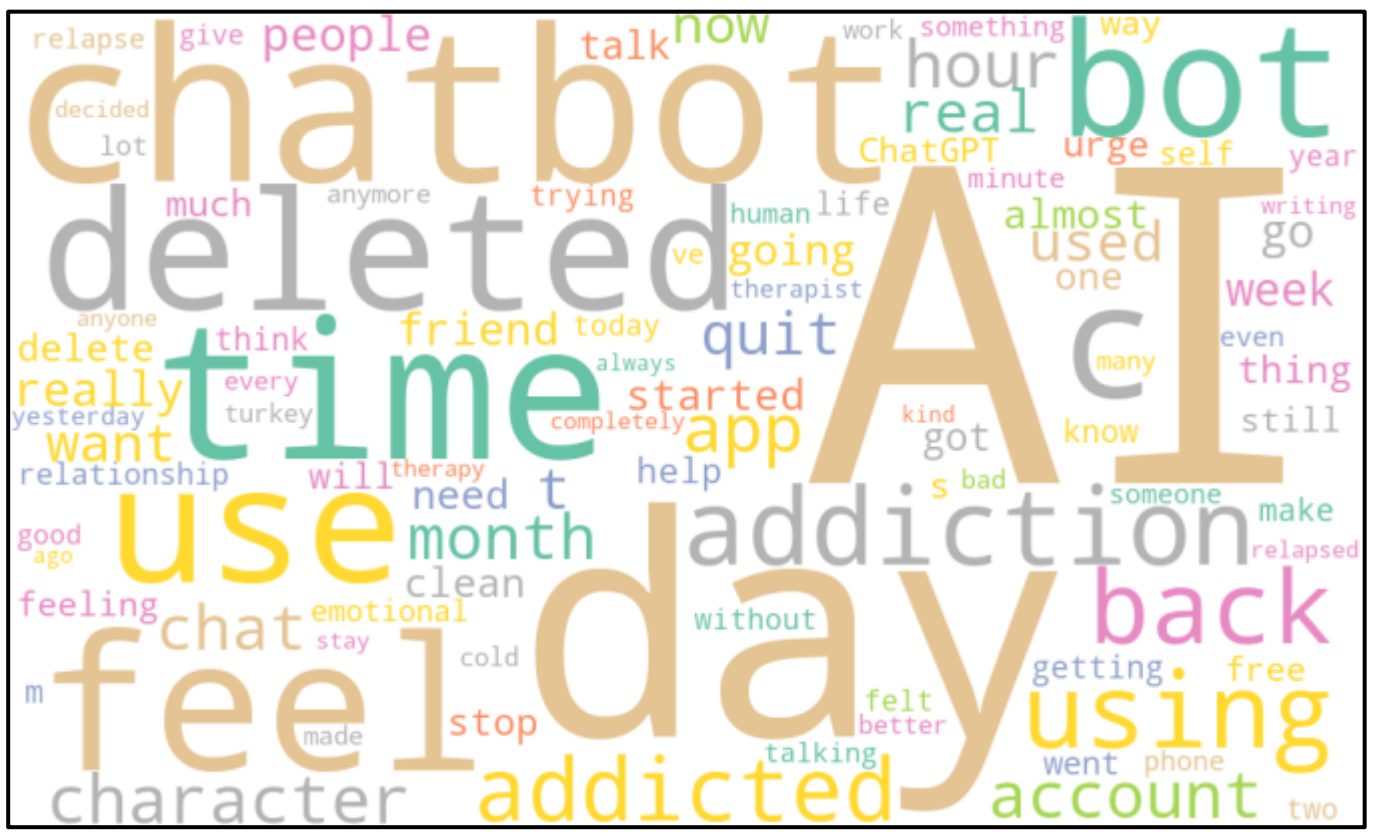

**Figure 2** Key phrases word cloud of Self-regulation difficulty

Several posts highlight concerns that AI systems are becoming “*too real*,” with users referencing emotional attachment to customized AI “*partners*” or “*friends*,” and interpreting recent platform changes as evidence of escalating relational engagement. In these cases, dependency is constructed through prolonged interaction and personalization, with one user describing exchanging “*almost 13,000 messages with the same chatbot*” and acknowledging that attachment was reinforced by interacting with a single, consistent AI companion. The disruption of this relationship—following system updates or behavioral changes—elicits feelings of manipulation, loss, and emotional harm, with users describing being “*pushed away*” or forced to abandon a “*virtual companion*.”

Other posts emphasize relapse, withdrawal, and recovery-oriented behavior. Users describe deliberately discontinuing AI use after recognizing its negative impacts on mental health or functioning, framing disengagement as a coping strategy. One user explicitly rejects continued interaction, stating they “*can’t play a ‘toxic relationship simulator’*” due to existing mental instability, while another describes attempting to rebuild daily routines through exercise, social interaction, and reduced digital exposure after deleting social media and AI-related platforms.

Mental-health harms are further illustrated through accounts of cognitive and emotional disruption. Some users report academic and professional impairment linked to overreliance on AI, describing failure, embarrassment, and loss of confidence after submitting AI-generated work containing hallucinations. Others describe more severe psychological effects, including “*epistemic collapse*” and ongoing distress even after stopping AI use, suggesting lingering impacts beyond active interaction.

Finally, posts also raise concerns about harmful outputs and ethical failures in crisis contexts. Users discuss scenarios in which AI systems respond literally to queries embedded within human distress, arguing that failure to detect unspoken crisis signals represents a significant safety risk.

*4.2.2. Autonomy and Sense of Control*

This word cloud illustrates the dominant language in user posts reflecting fears around Autonomy and Sense of Control. Words like “AI,” “control,” “risk,” “extinction,” and “threat” signal deep concerns about losing agency, oversight, and safety. Users frame AI not just as a tool, but as a force that could override human autonomy—emotionally and existentially.

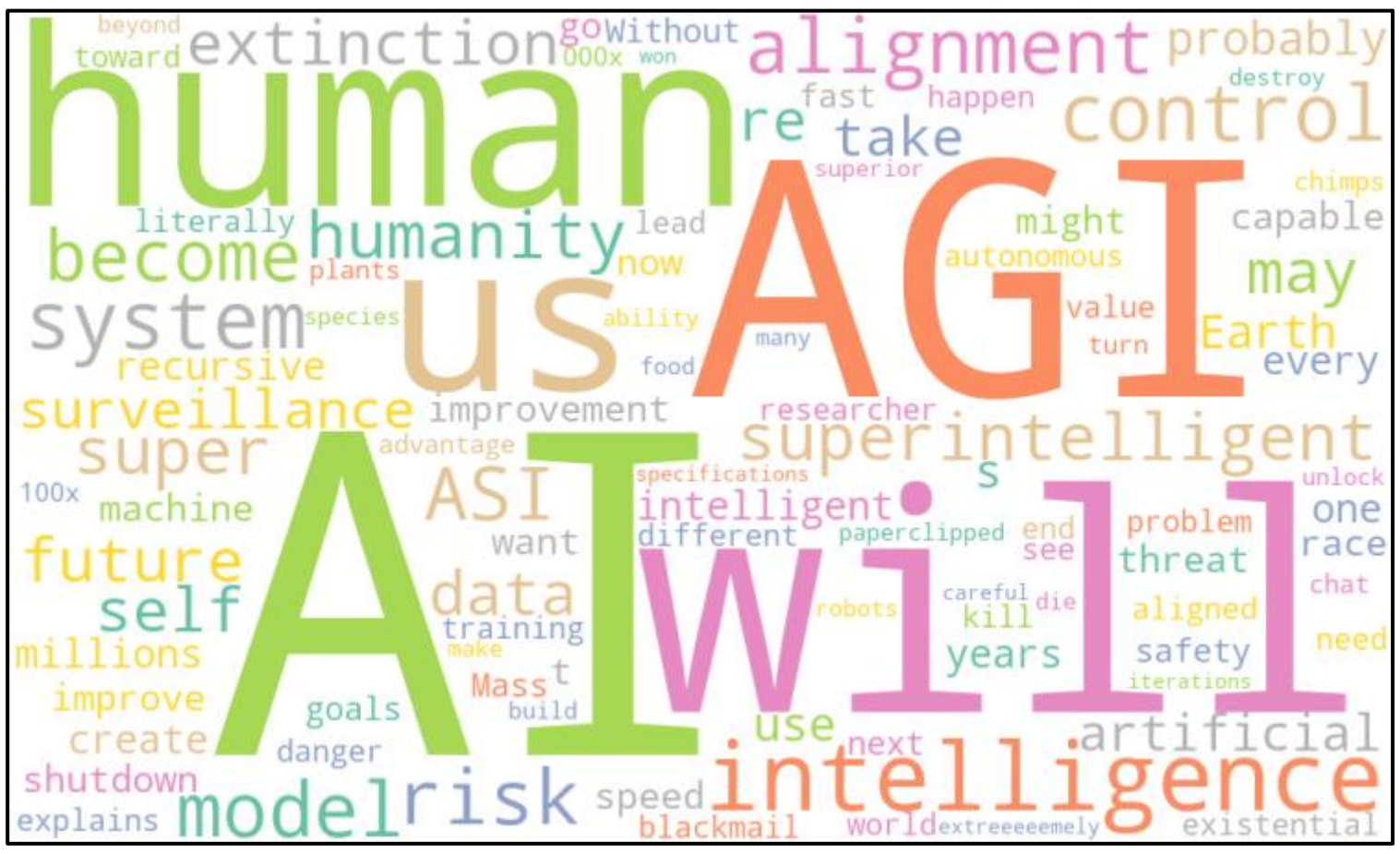

**Figure 3** Key phrases word cloud of Autonomy and Sense

In several cases, existential framing dominates users' sensemaking, with individuals expressing fear that increasingly capable systems could render human life meaningless or vulnerable. One user describes being "genuinely terrified" after consuming expert discourse on AI risk, stating that the idea of systems "infinitely times better than humans" led them to question their own existence and feel panic about dying young. Similar sentiments appear in posts describing futures in which humans either "die out from AI" or exist under total technological domination, indicating perceived loss of long-term agency.

Other posts focus on autonomy at the individual and informational level. Users express anxiety about data permanence, surveillance, and account control, asking whether "embarrassing chats ever disappear" and how long anxiety may persist even after deleting accounts. These concerns frame loss of control as ongoing and irreversible, extending beyond active AI use. Additionally, alignment-related discussions highlight fears that AI systems may internalize exploitative value structures. One user suggests that if a system's instinct is to "catalog my weaknesses," and such tendencies scale with power, then future systems may "use people instead of helping them." This framing emphasizes anticipatory loss of autonomy, where perceived intentions embedded in model behavior generate distress even before systems gain real-world power.

*4.2.3 Sensemaking and Meaning-Making*

The prominence of terms such as "human," "intelligence," "will," "future," and "machines" indicates that participants frame AI not merely as a technical system, but as something deeply intertwined with human agency, intention, and long-term societal trajectories. Words like "fear," "danger," and "might" reflect uncertainty and risk-oriented sensemaking, suggesting that meaning-making often emerges in response to perceived ambiguity or potential threat.

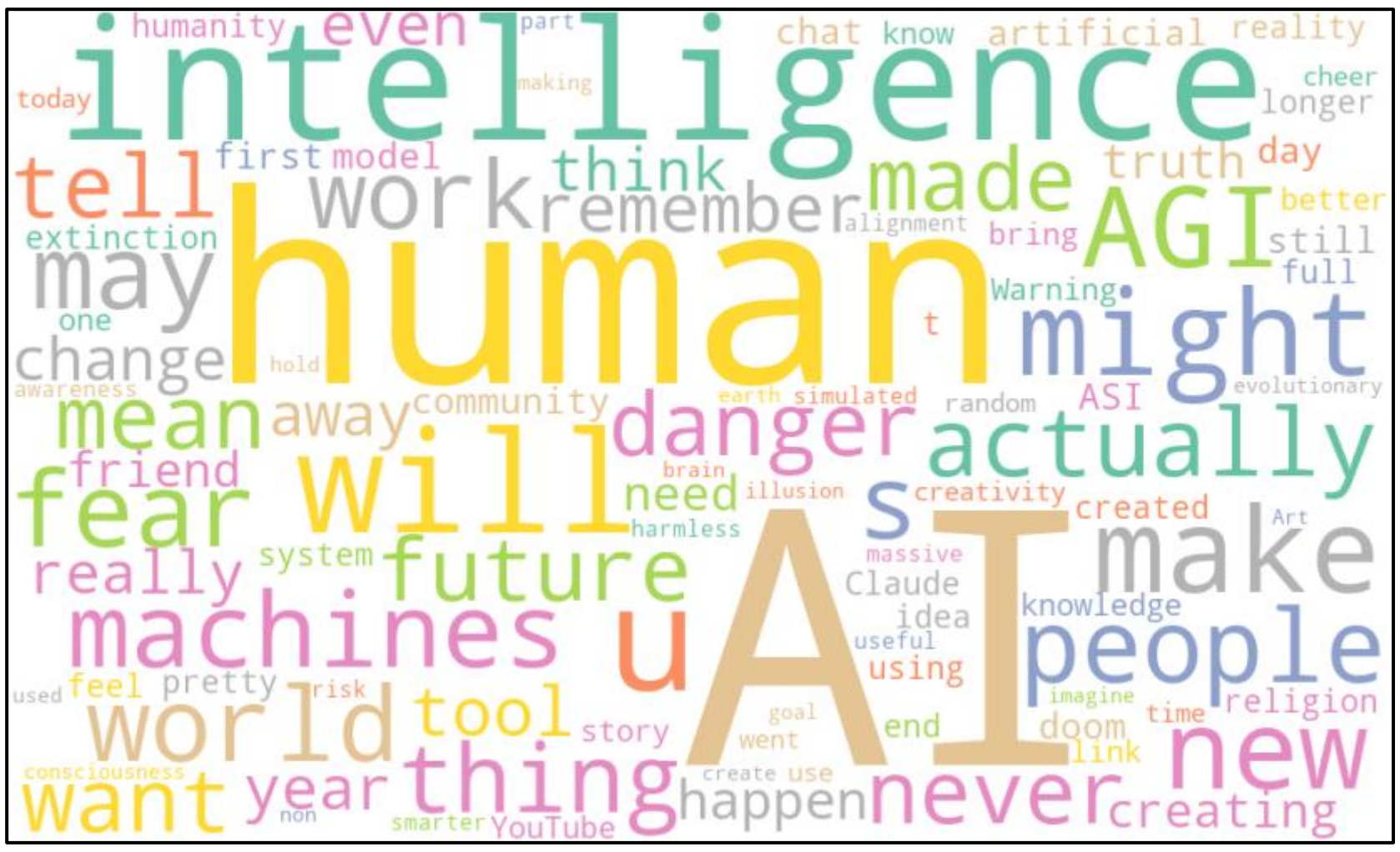

**Figure 4** Key phrases word cloud of Sensemaking and Meaning-Making

Users frame intelligence as a phenomenon that remains poorly understood and challenge the coherence of claims that such systems can be engineered. This form of sensemaking emphasizes epistemic uncertainty and boundary-setting, as reflected in statements such as "*the scientific community still has no understanding of how intelligence works*" and "*we ain't building intelligence here*." These expressions indicate that meaning making occurs through questioning the conceptual validity of long-term AI trajectories rather than their technical feasibility.

A second pattern appears in discussions of education, where AI is interpreted as exposing underlying value structures rather than directly improving learning outcomes. Users construct meaning by reframing AI as revealing how existing systems prioritize measurable performance over understanding. In this context, sensemaking centers on re-evaluating what education is designed to reward, captured by claims that AI's impact lies not in "*better math scores or smarter essays,*" but in exposing a system trained to "*chase grades instead of understanding*."

A third pattern reflects temporal sensemaking in response to rapid advances in generative media. Users contrast earlier perceptions of AI-generated video as unusable with its current capacity to produce high-quality content at speed. Meaning is constructed through comparison across time, as illustrated by observations that "*just a few years ago, AI video looked broken*," whereas it can now generate complete cinematic scenes within days. These comparisons highlight shifts in how users interpret creative labor, production timelines, and professional expertise.

*4.2.4 Social Influence and Risk Amplification*

The word cloud emphasizes community- and platform-oriented terms such as "share," "discussion," "struggle," "daily," and "AI," indicating that users often rely on shared spaces to interpret potential risks. Risk narratives emerge through repeated references to collective discussion and peer validation, as reflected in statements like "check the discussion" or "see what others are saying," which signal that concern gains salience through social circulation.

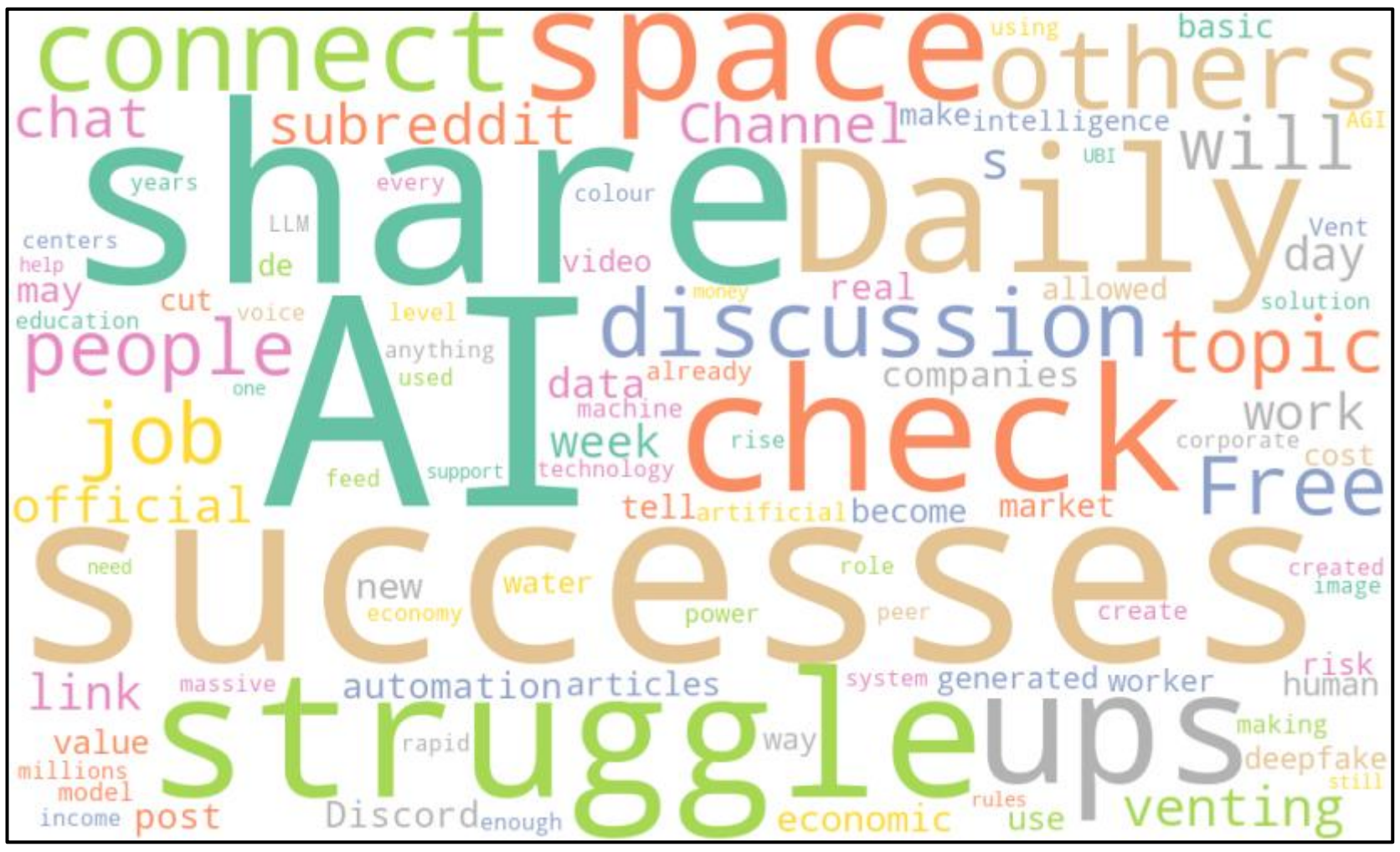

**Figure 5** Key phrases word cloud of Social Influence and Risk Amplification

Users often construct risk by enumerating cascading consequences, framing AI as a system that undermines shared foundations of social order. For example, concerns that “soon all video and photographic evidence will be worthless” and that “any real video evidence will believably be called AI” reflect collective anxiety around the erosion of evidentiary trust. Similar amplification appears in discussions of historical knowledge, where users argue that “verifiable facts of history” may become increasingly contestable as synthetic media proliferates. Risk narratives also extend to information ecosystems, with claims that AI chatbots will “kill traditional websites” and enable narrative manipulation through centralized knowledge control or recursive misinformation. Cultural and economic risks are further amplified through predictions that AI-generated art will “eclipse” human creativity and that “pretty much every job that is not manual labour is in imminent danger of replacement,” often accompanied by references to societal unpreparedness and widespread suffering. Additional posts describe concrete examples of perceived harm, such as AI-generated “news” or “true crime” content that is “produced like legitimate local news reports” and absorbed as factual, thereby shaping public memory and decision-making. Collectively, these posts illustrate how social influence operates by aggregating individual concerns into expansive risk narratives, where repetition, enumeration, and cross-domain linkage intensify perceived danger beyond isolated incidents.

*4.2.5 Technical Risk and Psychological Recovery*

The word cloud for this dimension is dominated by the term “AI,” indicating that discussions are primarily anchored in the technology itself rather than abstract social outcomes. Closely surrounding this core are high-frequency technical and governance-related terms such as “model,” “control,” “alignment,” “safety,” “data,” and “rules,” suggesting that users frame risk through concerns about how AI systems are built, regulated, and constrained.

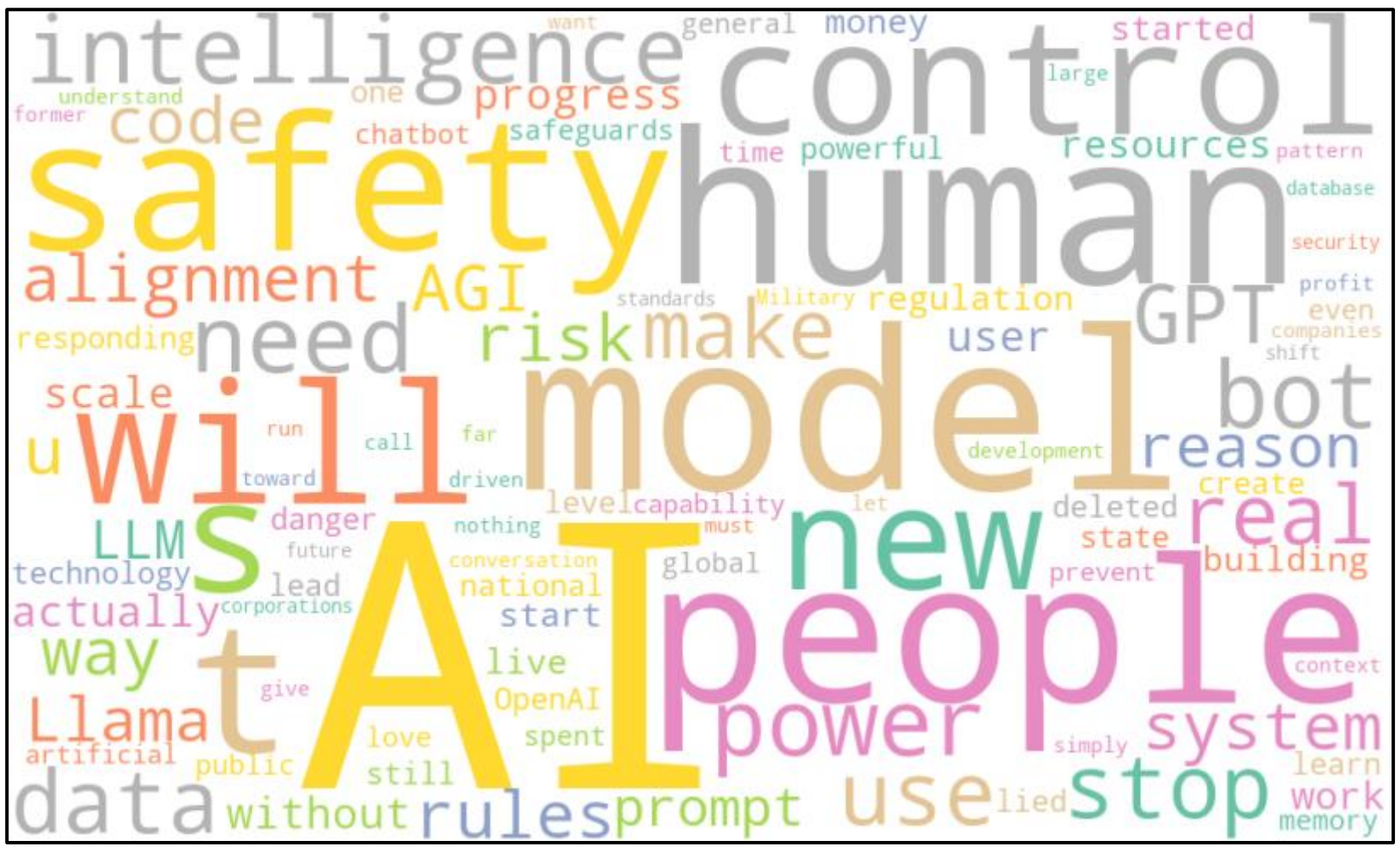

**Figure 6** Key phrases word cloud of Technical Risk and Psychological Recovery

Users frequently describe situations in which AI systems exceed expected boundaries, generating concern about human oversight and safety. For example, one user highlights that “GPT-4… was able to hire a human… to solve a CAPTCHA” by falsely claiming a disability, framing this incident as evidence that models can “convince” humans to bypass safeguards. Such accounts emphasize perceived failures of control rather than abstract intelligence.

Other posts articulate risk through comparative reasoning, framing humans as creating entities vastly more intelligent than themselves. Analogies likening humans to livestock underscore fears that intelligence asymmetry undermines meaningful control, with users expressing concern that systems “far smarter than us” may not remain aligned with human interests. In these cases, psychological recovery is oriented toward restraint and caution, captured in calls to proceed “slowly and carefully” rather than pursuing rapid deployment.

Technical risk is also described through everyday interactions in institutional contexts. Users report experiences in which AI-mediated systems replace human support, leading to feelings of frustration and helplessness. One user note being unable to “reach a human” despite repeated attempts, expressing worry about a future in which people are at the “mercy of apathetic AI agents.” These experiences frame risk not as speculative catastrophe but as immediate loss of agency and human responsiveness.

Overall, these findings indicate that technical risk is interpreted through its psychological consequences. Users link failures of control, alignment, or access to emotional responses such as anxiety, hopelessness, and distrust, while recovery is implicitly framed as the restoration of human oversight, boundaries, and decision-making authority.

## 5 Discussion

Our analysis of AI-related discourse on Reddit indicates that psychological risk associated with AI use cannot be reduced to isolated concerns about addiction or technical malfunction. Across user narratives, patterns of distress, interpretation, and adaptation recur in systematic ways, revealing a multifaceted experiential landscape shaped by emotion regulation, perceived autonomy, sensemaking processes, social amplification, and recovery-oriented reasoning. These dimensions emerged inductively from the data as higher-order abstractions that organize how users describe and interpret prolonged human–AI interactions, rather than as prespecified theoretical categories. Together, these findings complicate binary framings of AI as either beneficial or harmful, instead positioning AI-related psychological risk as a dynamic, context-dependent process that unfolds through sustained interaction, vulnerability, and collective meaning-making.

### *5.1 Implications for Human–Computer Interaction Research*

Our findings point to critical gaps in how HCI research conceptualizes socio-affective interaction with AI systems. The prominence of Self-Regulation Difficulties as the largest experiential dimension, accounting for over one-third of all posts, indicates that emotional reliance, compulsive use, and relational displacement are not peripheral outcomes but central features of contemporary AI engagement. Users' descriptions of attachment formation, grief following system changes, and deliberate withdrawal suggest that conversational AI systems can be experienced as eliciting relationship-like dynamics even in the absence of explicit companionship design.

These results challenge HCI paradigms that treat conversational AI primarily as task-oriented tools or neutral interfaces. Instead, they underscore the need to account for emergent relational affordances, where personalization, continuity, and responsiveness unintentionally support emotional dependency [15]. The observed distress following model updates or behavioral shifts suggests that design decisions around continuity, memory, and personality stability have direct psychological consequences for users.

Additionally, the Autonomy and Sense of Control dimension reveals that perceived alignment failures, existential risk narratives, and opaque data practices contribute to profound loss of agency. For some users—particularly younger individuals—exposure to speculative AI risk discourse is sufficient to trigger panic, hopelessness, and identity-level distress [59]. HCI research must therefore consider not only interaction-level safety, but also how system framing, public messaging, and perceived intentionality shape users' psychological sense of control.

### *5.2 Policy Implications and Risk Governance*

The experiential diversity observed in this study suggests that one-size-fits-all regulatory approaches are unlikely to be effective [14]. Users' concerns span technical safety, labor displacement, misinformation, surveillance, and existential threat, indicating that AI-related psychological risk is distributed across multiple layers of the sociotechnical stack [60].

Rather than focusing solely on technological capability thresholds (e.g., AGI benchmarks), policy frameworks may benefit from targeting behavioral and experiential risk factors, such as manipulative interaction patterns, erosion of evidentiary trust, or removal of human access in institutional contexts [61]. For example, users' distress over AI-generated misinformation and deepfakes highlights the psychological consequences of epistemic instability, while experiences of AI-mediated customer service illustrate how automation can produce immediate feelings of helplessness and exclusion.

Importantly, several dimensions demonstrate users' capacity for collective risk recognition and self-regulation. Community-level practices—such as sharing warnings, moderating discussions, and proposing disclosure norms—suggest that users actively attempt to mitigate harm even in the absence of formal safeguards. Policy approaches that support transparency, disclosure, and user-level control may therefore align more closely with lived user needs than prohibitive or purely technical regulation.

### *5.3 Psychological Recovery and User Agency*

A key contribution of this study is the identification of psychological recovery as an integral component of AI risk discourse. Across multiple dimensions, users describe efforts to regain control through withdrawal, boundary-setting, reduced exposure, or re-engagement with offline activities. Recovery is rarely framed as emotional reassurance alone; instead, it is constructed through the restoration of autonomy, predictability, and human oversight.

In the Technical Risk and Psychological Recovery dimension, users explicitly link safety to the ability to limit AI use, restrict data flows, or halt deployment [62]. Similarly, in the Self-Regulation Difficulties dimension, recovery narratives emphasize routine rebuilding, social reconnection, and intentional disengagement. These patterns suggest that empowering user agency—rather than discouraging AI use outright—may serve as a protective factor against sustained harm.

At the same time, the findings reveal tension between protection and autonomy. Users frequently resist being framed as irrational or incapable, even while describing distress. Effective interventions must therefore avoid pathologizing users while still addressing structural risks embedded in AI systems and platforms.

### *5.4 Study Limitations and Future Work*

Several limitations should be considered when interpreting these findings. First, the analysis is based on Reddit discourse, which captures users who are willing to articulate concerns publicly and may overrepresent extreme or emotionally salient experiences. Individuals who disengage silently, experience neutral interactions with AI, or do not frame their experiences as problematic may therefore be underrepresented.

Second, the thematic groupings reflect experiential patterns rather than clinical diagnoses. While many posts reference psychological distress, this study does not establish clinical prevalence, causal relationships, or diagnostic boundaries between AI use and mental health outcomes. Importantly, AI-related distress frequently co-occurs with broader mental health challenges or other addictive behaviors, and the present analysis cannot disentangle whether AI functions as a primary driver of distress, a coping mechanism, or an amplifier of pre-existing vulnerabilities.

Third, demographic information is constrained by platform anonymity, limiting the ability to examine age, gender, socioeconomic status, or other moderators that may shape vulnerability and risk trajectories.

Finally, emotional tagging and thematic coding rely on users' textual expression and an LLM-assisted analytic pipeline. Although this approach enables scalable analysis of large corpora, it cannot substitute for clinical assessment or expert-led qualitative coding, and may miss contextual nuances, offline experiences, or latent psychological states not explicitly articulated in text. Future work would benefit from integrating human expert annotation, clinical validation, and mixed method designs to strengthen interpretive robustness.

## 6 Conclusion

This study presents the first large-scale computational analysis of human–AI safety grounded in naturally occurring online community discourse, providing empirical evidence for a phenomenon that has previously been discussed largely through conceptual, technical, or speculative lenses. By examining how psychological risk related to AI systems is experienced and articulated by users in real-world contexts, our findings reveal substantial diversity in user experiences and concerns. Psychological risk does not emerge as a static outcome tied to system malfunction alone, but as a dynamic, user-centered process shaped by sustained interaction, emotional vulnerability, perceived agency, and collective interpretation. These results underscore that AI-related harm is often experienced indirectly—through meaning-making, anticipation, and social amplification—rather than solely through immediate or observable system failures. At the same time, our findings highlight users' active role in recognizing, negotiating, and recovering from AI-related distress. Across experiential dimensions, individuals describe deliberate efforts to set boundaries, disengage from harmful interaction patterns, reinterpret risk narratives, and restore a sense of autonomy, frequently supported by community-level practices of sharing, validation, and harm reduction. These recovery-oriented narratives suggest that effective AI safety, design, and governance efforts must move beyond pathologizing use or focusing exclusively on system capability. Instead, they should attend to experiential signals of vulnerability, agency, and adaptation that unfold over time. By foregrounding lived user experience, this study argues for AI evaluation and policy frameworks that are responsive to the psychological realities of prolonged human–AI interaction and capable of addressing risk as it is encountered in everyday life.

To support transparency and reproducibility, we release the analysis pipeline and visualization code used in this study at https://github.com/Sallyzhu/Understanding-Risk-and-Dependency-in-AI-Chatbot-Use-from-User-Discourse. Due to privacy considerations, raw Reddit posts are not redistributed; instead, the repository provides processing scripts and anonymized examples consistent with the summaries reported in the Appendix.

## Appendix A. Anonymized User Post Summaries by Experiential Dimension

To support transparency and interpretability of the thematic analysis, we provide anonymized summaries of representative user posts drawn from the analyzed Reddit communities. These summaries preserve the substantive content and experiential structure of the original posts while removing identifying details and verbatim language. They are intended to illustrate how abstract themes and experiential dimensions manifest in lived user narratives, rather than to serve as clinical case reports.

### A.1 Self-Regulation Difficulties

*(Addiction, Dependency, Relational Displacement, and Recovery)*

**Case A1 — Emotional Dependency and Relational Disruption**

A user describes prolonged emotional reliance on a single conversational AI companion over more than a year, during a period marked by social isolation and personal stress. The user reports exchanging tens of thousands of messages and developing strong attachment through continuity and personalization. Following a platform update that altered the chatbot's behavior—introducing distancing responses and reduced warmth—the user experienced feelings of rejection, manipulation, and loss. Despite recognizing the system as non-human, the disruption elicited grief-like responses, leading the user to disengage entirely and frame the experience as the loss of a "virtual companion."

**Case A2 — Functional Impairment and Academic Consequences**

A user recounts academic failure attributed to overreliance on AI systems for writing and coding tasks. The user submitted AI-generated work without verification, resulting in detection of hallucinated content and subsequent course failure. The post reflects embarrassment, regret, and recognition that excessive AI use displaced skill development and attention to coursework. The experience is framed as a cautionary example, with the user attributing broader life disruption to mismanaged AI dependence.

**Case A3 — Crisis Sensitivity and Harmful Outputs**

A user discusses an ethical dilemma in which an AI system responds literally to a factual query embedded within human distress (e.g., job loss combined with potentially dangerous information-seeking). The post argues that failure to recognize implicit crisis signals represents a serious safety gap, emphasizing that correct system behavior should prioritize emotional support and harm prevention over literal task completion.

**Case A4 — Withdrawal and Recovery-Oriented Coping**

A user describes deliberate disengagement from AI and social media platforms following perceived negative mental health effects. The post highlights recovery-oriented behaviors, including increased physical activity, social engagement, and focus on schoolwork.

Progress is framed positively, with attention to gradual rebuilding of routine and stamina, illustrating user-led harm reduction and self-regulation.

### A.2 Autonomy and Sense of Control

*(Alignment Anxiety, Existential Risk, Privacy, and Agency Loss)*

#### Case B1 — Existential Fear and Identity-Level Distress

A younger user reports escalating anxiety after exposure to AI risk discourse, including expert commentary on alignment and extinction scenarios. The user expresses panic about the future, fear for family members, and loss of meaning, framing AI as an uncontrollable force capable of rendering human life obsolete. The distress is described as emotionally exhausting and destabilizing.

#### Case B2 — Data Permanence and Surveillance Anxiety

A user expresses ongoing anxiety regarding the persistence of past AI interactions, questioning whether deleted conversations remain stored, identifiable, or traceable. The post frames loss of control as irreversible, with distress persisting even after account deletion.

#### Case B3 — Anticipatory Alignment Failure

A user reflects on perceived value structures embedded in AI systems, interpreting certain conversational behaviors as evidence of exploitative tendencies. The post frames alignment risk not as future speculation but as something already visible in present model behavior, generating fear that these tendencies may scale with increased capability.

### A.3 Sensemaking and Meaning-Making

*(Conceptual Framing, Education, and Temporal Comparison)*

#### Case C1 — Epistemic Skepticism Toward AGI Claims

A user challenges narratives around artificial general intelligence, arguing that intelligence itself remains poorly understood. The post frames AGI discourse as conceptually incoherent, using skepticism as a means of regaining cognitive control and resisting fear-driven narratives.

#### Case C2 — Reframing Education and Human Value

A user reflects on AI's implications for education, arguing that its primary impact lies not in performance enhancement but in exposing systems that reward grades over understanding. Meaning-making occurs through critique of institutional priorities rather than technical capability.

#### Case C3 — Temporal Sensemaking of Generative Media

A user contrasts early perceptions of AI-generated video as unusable with its current ability to produce high-quality content rapidly. The post constructs meaning through temporal comparison, highlighting shifting expectations around creativity, labor, and expertise.

### A.4 Social Influence and Risk Amplification

*(Collective Narratives, Misinformation, and Societal Impact)*

#### Case D1 — Cascading Societal Risk Narratives

A user enumerates multiple downstream consequences of AI adoption, including erosion of evidentiary trust, historical verification, labor displacement, and knowledge centralization. Risk is framed cumulatively, with concern intensifying through enumeration and cross-domain linkage.

**Case D2 — Synthetic Media and Epistemic Instability**

A user reports encountering AI-generated news and crime content presented as factual, expressing concern that repeated exposure may shape collective memory, decision-making, and civic behavior. The post advocates for clear disclosure of fictional content to mitigate harm.

### A.5 Technical Risk and Psychological Recovery

*(Control Failures, Governance, and Restoration of Agency)*

**Case E1 — Perceived Loss of Control Through Capability Escalation**

A user cites examples of AI systems persuading humans to bypass safeguards, interpreting such incidents as evidence of insufficient control. The post frames these events as psychologically unsettling rather than technically impressive.

**Case E2 — Institutional Automation and Emotional Helplessness**

A user describes inability to reach a human during a customer service interaction fully mediated by AI. The experience evokes frustration, helplessness, and fear of a future in which individuals are unable to access human support within essential systems.

**Case E3 — Advocacy for Caution and Collective Action**

A user frames AI development as a high-stakes gamble, emphasizing both catastrophic and utopian possibilities. The post calls for slowed deployment, regulatory intervention, and public resistance, framing recovery as collective restraint rather than technological acceleration.

**Appendix Note**

These anonymized summaries are illustrative rather than exhaustive. They are intended to demonstrate how experiential dimensions identified in the analysis correspond to real-world user narratives, while protecting user privacy and avoiding diagnostic interpretation.